\documentclass[11pt]{scrartcl}
\setkomafont{disposition}{\normalcolor\bfseries}
\usepackage[charter,greeklowercase=upright]{mathdesign}

\usepackage{graphicx}
\PassOptionsToPackage{british}{babel}  
	\usepackage{babel}                  

\usepackage{tabu}
\usepackage{tabularx}
\usepackage{booktabs}
\usepackage{url}
\usepackage{hyperref}
\usepackage{microtype}
\usepackage{ntheorem}

\usepackage[threshold= 1]{csquotes}

\PassOptionsToPackage{
	backend=biber,
	bibencoding=ascii,%
	language=auto,
    style= apa,
    natbib=true
}{biblatex}
    \usepackage{biblatex}

\DeclareLanguageMapping{british}{british-apa}
\setkomafont{author}{\normalsize}
\setkomafont{date}{\normalsize}
\title{\LARGE Enslaving the Algorithm: From a ``Right to an Explanation'' to a ``Right to Better Decisions''?}

\author{Lilian Edwards, University of Strathclyde [l.edwards@strath.ac.uk]\\ Michael Veale, University College London [m.veale@ucl.ac.uk]}

\date{\textit{Published in}\\ IEEE Security \& Privacy (2018) 16(3), 46--54, doi:\href{https://doi.org/10.1109/MSP.2018.2701152}{10.1109/MSP.2018.2701152}}

\begin{document}
\sloppy
\maketitle

\begin{abstract}
 As concerns about unfairness and discrimination in ``black box'' machine learning systems rise, a legal ``right to an explanation'' has emerged as a compellingly attractive approach for challenge and redress. We outline recent debates on the limited provisions in European data protection law, and introduce and analyze newer explanation rights in French administrative law and the draft modernized Council of Europe Convention 108. While individual rights can be useful, in privacy law they have historically unreasonably burdened the average data subject. ``Meaningful information'' about algorithmic logics is more technically possible than commonly thought, but this exacerbates a new ``transparency fallacy''---an illusion of remedy rather than anything substantively helpful. While rights-based approaches deserve a firm place in the toolbox, other forms of governance, such as impact assessments, ``soft law,'' judicial review, and model repositories deserve more attention, alongside catalyzing agencies acting for users to control algorithmic system design.\end{abstract}

\section{Introduction}

Businesses and governments are increasingly deploying machine learning (ML) systems to make and support decisions that have a crucial impact on everyday life: decisions about (inter alia) criminal sentencing and release on bail, medical treatment, eligibility for welfare benefits, what entertainment we see and can access, the price and availability of goods and services delivered online, and the political information to which we are exposed. These ML systems---colloquially entering public consciousness as just \emph{algorithms}, or even just \emph{AI}---have been extensively criticized in the past few years as a result of a number of well-known ``war stories'' that have revealed patterns of discrimination embedded but invisible to casual users in such systems.\textsuperscript{1}

Because algorithms are trained on historical data, they risk replicating unwanted historical patterns of unfairness and/or discrimination. For example, in hiring systems, a lack of women being hired in the past may mean the systems fail to recognize the worth of female applicants, or even outright discriminate against them. Luxury goods may be advertised to people with certain profiles on social media and not to others, creating a consumer ``under class.''

A severe obstacle to challenging such systems is that outputs, which translate with or without human intervention to decisions, are made not by humans or even human-legible rules, but by less scrutable mathematical techniques. A loan applicant denied credit by a credit-scoring ML algorithm cannot easily understand if her data was wrongly entered, or what she can do to have a greater chance of acceptance in the future, let alone prove the system is illegally discriminating against her (perhaps based on race, sex, or age). This opacity has been described as creating a ``black box'' society.\textsuperscript{2}

\section{Enter the Right to an Explanation}\label{enter-the-right-to-an-explanation}

Since the 1990s, the law in Europe has been concerned with this kind of opaque and difficult-to-challenge decision making by automated systems. In consequence, the Data Protection Directive (DPD), a measure that harmonized relevant law across EU member states in 1995, provided that a ``significant'' decision could not be on based solely on automated data processing (article 15). Some EU members interpreted this as a strict prohibition, others as giving citizens a right to challenge such a decision and ask for a ``human in the loop.'' A second right, embedded within article 12, which generally gives users rights to obtain information about whether and how their particular personal data was processed, gave users the specific right to obtain ``knowledge of the logic involved in any automatic processing'' of their data. Both these provisions, but especially the latter, were not much noticed, even by lawyers, and scarcely ever litigated, but have revived in significance in the latest iteration of EU data protection (DP) law within the General Data Protection Regulation (GDPR), which passed in 2016 and will come into operation across Europe in 2018.

In the GDPR, article 15 has been transformed into Article 22 and has arguably created what the media and some technical press have portrayed as a new ``right to an explanation'' of algorithms. The former article 12 has also been revamped to a new article 15 and now includes a right to access to ``meaningful information about the logic involved, as well as the significance and the envisaged consequences of such processing'' (article 15(1)(h)). This provision, notably, applies only in the context of ``automated decision making in the context of'' Article 22. This leaves it unclear if all the constraints on Article 22 (discussed below) are ported into article 15 (though our view is that it does not). Sadly, all this adds up to a reality considerably foggier than the media portrayal.

Several factors undermine the idea that Article 22 contains a right to an explanation.

Primarily, Article 22 does not in its main thrust even contain a right to an explanation, but is merely a right to stop processing unless a human is introduced to review the decision on challenge. However, Article 22 does refer at points to a requirement of ``safeguards,'' both where the right to prevent processing (paradoxically) does not operate, and where it does but sensitive personal data is processed. In relation to the first case, safeguards are partly listed in Article 22(3), but in the second case, the only guidance is in Recital 71. (``Sensitive'' personal data in DP law refers to a restricted list of factors regarded as particularly important such as health, race, sex, sexuality, and religious beliefs.)

It is important to note that, in European legislation, the articles in the main text are binding on member states but are accompanied by ``recitals,'' which are designed to help states interpret the articles and understand their purpose. Recitals are usually regarded as helpful rather than binding, but this is contested and differs among states. Unfortunately, in relation to Article 22, Recital 71 mentions some key matters not included in the main text. Article 22(3) mandates that safeguards include ``at least the right to obtain human intervention on the part of the controller, to express his or her point of view and to contest the decision,'' but the safeguards listed in Recital 71 ``should include specific information to the data subject and the right to \emph{obtain human intervention, to express his or her point of view, to obtain an explanation of the decision reached after such assessment} and to challenge the decision'' (italics added).

This strange mishmash of texts thus cannot firmly be said to mandate a right to explanation in all or indeed any circumstances and may not be interpreted the same way from state to state.

This is a serious, but not the only, problem with Article 22.

\begin{itemize}
\item
  Article 22 applies only to systems where decisions are made in a ``solely'' automated way---that is, there is no human in the loop---and there are very few of these and fewer that are ``significant'' (see below). How meaningful this input has to be is subject to recent regulatory guidance,\textsuperscript{3} but remains unclear and untested.
\item
  What is a ``decision''? The GDPR gives us no help with this at all other than that it includes a ``measure'' (Recital 71). Is sending a targeted ad to a user using an algorithmic system a decision? It produces no binding effect; the advert may be ignored; and in many cases, it is hard to see what action causally flows from it. Yet as in the well-publicized Latanya Sweeney example,\textsuperscript{4} sending adverts promoting help with criminal arrests solely to ``black-sounding'' names was worrying and offensive---and potentially dangerous if these characterizations were inherited by systems selecting individuals for stop and search or airport screening. Although a single advert delivery decision might not have a significant effect on an individual's life, the cumulative effect on an entire group or class may be worrying. Such group privacy impacts are not dealt with well by DP law---an area based on individualistic human rights---and are exacerbated by a continuing lack of provision for class actions in EU states.
\item
  Article 22 applies only to a decision that produces legal or other ``significant'' effects. This is vague in the extreme. Some would argue this could only apply to systems that make important, binding decisions on things like criminal justice, risk assessment, credit scoring, education applications, or employment. Yet such systems are rarely if ever entirely automated, even if the human's involvement is often nominal. Furthermore, some commercial decisions may seem trivial as a one-off, but are significant in aggregate. Mendoza and Bygrave argue that advertising decisions can never be significant,\textsuperscript{5} while European regulators recently produced guidance indicating the opposite.\textsuperscript{3} Might systems recommending buying choices or targeting adverts not limit a user's worldview or choices, or disseminate ``fake news'' via algorithmic filter bubbles? Arguably, such phenomena are becoming deeply and significantly destructive to our democracy. We have an obvious link here to the issue of to whom a decision needs to be significant: the individual in question or society as a whole?
\end{itemize}

Turning to the new article 15 of the GDPR (right to information), this right to ``meaningful information about the logic involved'' in any decision-making system may be more useful than Article 22. It is (arguably) not directly as restrictive as Article 22 is to solely automated decisions having significant or legal effects. But there is an unresolved doubt about whether it only applies to information available before the system makes a decision about a particular data subject (see Wachter and colleagues' work\textsuperscript{6}). In ML parlance, that means it is uncertain if the right is only to a general explanation of the model of the system as a whole (model-based explanation), rather than an explanation of how a decision was made based on that particular data subject's particular facts (subject-based explanation).\textsuperscript{1}

But even if we agree that Article 15 may give us some kind of functioning right to an explanation, we still have huge problems. The GDPR can apply only where decisions are made based on personal data. Personal data is defined in article 4(1), as ``any information relating to an identified or identifiable natural person'' and is certainly wider than data that has the name of a data subject attached to it. According to Recital 26:

To determine whether a natural person is identifiable, account should be taken of all the means reasonably likely to be used, such as singling out, either by the controller or by another person to identify the natural person directly or indirectly.

Even allowing for this broad (and still controversial) approach to defining personal data, some data will remain outside the scope of the GDPR. Yet algorithmic decisions that affect people may not involve personal data. Take self-driving cars. They may kill people---passengers or pedestrians---as a result of algorithmic processing, yet the data involved in that decision may be entirely related to traffic, road conditions, and other non-personal matters. Other circumstances may involve data that was once personal but has been allegedly anonymized. This is very common, for example, with profiles made from personal data collected by social networks and used to generate targeted marketing.

Lastly, a final restriction to these rights comes in the form of a carve-out in recital 63 (though not main text) for intellectual property (IP) and trade secrets. Explanations of how an algorithm works might reveal a firm's competitive advantage---its notorious ``secret sauce.'' Although this should not result in a ``refusal to provide all information to the data subject,'' the lack of clear-cut provisions will likely continue to further muddy the waters. In these cases, proposed explanations of systems based on modeling slices of a proprietary model without access to its innards may prove a useful middle ground.\textsuperscript{7,8}

\subsection{Improving the Right to an Explanation after the GDPR}\label{improving-the-right-to-an-explanation-after-the-gdpr}

Above, we have seen how the GDPR's right to an explanation appears far from ideal. This is hardly surprising given that the provisions remain largely modeled on the 1995 directive, which effectively predated the Internet and modern algorithm design and accompanying harms.

More modern laws exist: one example is the French \emph{loi pour une République numérique} (Digital Republic Act, law no. 2016-1321), This gives a right to an explanation for \emph{administrative} algorithmic decisions made about individuals. The new law provides that in the case of ``a decision taken on the basis of an algorithmic treatment'' (author translation), the rules that define that treatment and its ``principal characteristics'' must be communicated upon request. Further details were added by decree in March 2017 (R311-3-1-2) elaborating that the administration shall provide information about:

\begin{enumerate}
\item the degree and the mode of contribution of the algorithmic processing to the decision making;
\item the data processed and its source;
\item the treatment parameters and, where appropriate, their weighting, applied to the situation of the person concerned; and
\item  the operations carried out by the treatment.
\end{enumerate}

Some areas of government, such as national security and defense, are excluded.

The French approach has important advantages over the GDPR. First, considering point 1, true decision-support systems are not automatically excluded from being explained.

Secondly, point 3 provides that, where appropriate, the weightings of factors in a system can be disclosed. This seems to imply the explanation must be of a particular decision (subject-based explanation) rather than a vague overview of a complex model (model-based explanation), in contrast to the reading of the GDPR by Wachter and colleagues.\textsuperscript{6} Extracting estimates of the weightings within a complex algorithm is increasingly possible, particularly if only the area ``local'' to the query is being considered,\textsuperscript{7} which unlike the complex innards of the entire network, might display recognizable patterns.\textsuperscript{9}

Weights may help explain systems, but they are by no means a complete fast track to interpretability. There are at least two occasions when a court might say that weights are not useful for explaining a decision to a human user, and therefore, it is not appropriate to order disclosure. These are when the weighted inputs do not map to any real-world features the user will find intelligible and, in older or restricted systems, where retrofitting an explanation system is infeasible.

On the downside, the new French right applies only to administrative decisions. This makes its advances more comprehensible, for several reasons. First, the number of discretionary decisions currently made by governmental algorithmic systems is relatively small compared to the increasing amount of ML profiling in the private/commercial sector. Second, there is a long-established constitutional expectation that democratic governments will, to some extent, be transparent---via, for example, freedom of information requests. By contrast, the private sector is usually not required to disclose its secrets except on limited occasions such as financial disclosures.

Another new instrument in the planning is the modernization of the Council of Europe Convention 108 for the Protection of Individuals with Regard to Automatic Processing of Personal Data (CoE 108). CoE 108 is an international treaty relating to DP, which regardless of its name, can be signed by any state in the world. Its membership, however, remains relatively limited and, for instance, does not currently include either the US or China.

In a recently circulated draft, it appeared that the CoE was considering as one alternative, a version of the right to an explanation for all automated decisions, without the restriction of GDPR Article 22. This would be an exciting development. While the outcomes of the CoE negotiations are far from settled, interestingly it seems this potentially expanded CoE right has already been ported to the draft UK transposition of the GDPR, the Data Protection Bill 2017 (see https://perma.cc/X7X6-TXW9). The bill is oddly drafted but, in brief, the first two parts of the bill reflect EU law since they transpose both the GDPR and the new EU directive relating to DP in policing matters (Directive (EU) 2016/680). The third part of the bill, however, applies DP law to UK intelligence services. These are outside of EU law, but the UK has chosen to make these provisions compliant not with the GDPR but with CoE 108. As a result, the draft bill's s96 currently contains one of the most advanced provisions on a right to explanation in the world! Before too much excitement is generated, however, it should be noted that any disclosures will still be controlled by overarching exemptions for national security in draft s108 and s109, and so are in fact never likely to be exercised.

\section{Is a Right to an Explanation Our Best Remedy?}\label{is-a-right-to-an-explanation-our-best-remedy}

The right to an explanation is only one tool for scrutinizing, challenging, and restraining algorithmic decision making. While it has rhetorical strength in demanding transparency to enable user challenge, it has serious practical and conceptual flaws.

First, reliance on an explanation to bolster individual rights places a primary and heavy onus on users to challenge bad decisions. Even ordinary DP subject access requests (SARs) demand an enormous amount of time and persistence and, in reality, are mainly used effectively only by journalists and insiders who know how the company in question organizes its data processing systems. Very few ordinary users historically make use of SARs, and still fewer will probably use the right to an explanation. These issues are abetted by the common problems of consumer access to justice, including a general lack of access to legal aid and, in the EU, to class actions or collective redress. As noted below, users are supposed to be represented by their state data protection authorities (DPAs) in Europe---but in reality, this support may be lacking due to a dearth of personpower and expertise in these bodies.

Second, even if obtained, an explanation may not be helpful in mounting a challenge. This may not be just because of the well-known difficulties about expressing machine logic in human-comprehensible form,\textsuperscript{1} despite the progress that is being made in overcoming the hurdles to producing ``meaningful information'' about algorithmic logic. But algorithmic models, inputs, and weightings, however disclosed, may still not show that a system has been designed to be biased, unfair, or deceptive. Most algorithms will display inadvertent bias rather than explicitly coded-in bias. (Designers will not want to be sued or prosecuted for illegal action even if their own ethics do not forbid it.) Researchers are discovering now how difficult many of these problematic but non-obvious issues can be to spot even when they have the whole dataset to hand. Biased or discriminatory behavior may only become apparent looking at the corpus of users as a whole---something that will not happen through individual user challenges.

It might be possible to better understand these aspects of a model as a whole if many individuals could utilize their individual rights to explanation at once. This would require far better legal and technical mechanisms for collective action and challenge than we have now. At the moment, even gathering information about the collective impact of algorithmic systems on users is difficult and unusual. A good example is the sending of ``dark ads'' to voters during recent political campaigns via political profiling of voters on social media platforms. Because these adverts were personally targeted to users, outside agencies could not even know what ads were being deployed, less still count them or their influence. Dark ads during the 2016 British general election were tracked to some extent by volunteers who installed and browsed the Internet with the WhoTargetsMe tool, but naturally this counting was very partial and non-representative. Given the already weighty burden on the individual, also requiring them to coordinate collective action to expose unfair algorithms seems unlikely to succeed.

In short, a legal right to an explanation may be a good place to start, but it is by no means the end of the story. Rights become dangerous things if they are unreasonably hard to exercise or ineffective in results, because they give the illusion that something has been done while in fact things are no better. It is instructive here to compare the history of consent to sharing of data, which has moved in the online world from a real bulwark of privacy to something most often described as meaningless or illusory. Consent is usually given via privacy policies that are largely never read---and, if read, not understood---and they cannot be negotiated and change from time to time. Thus, consent has become a formality validating the actions of the data controller rather than something empowering the user. This is sometimes known as the \emph{notice and choice fallacy}. It would be worrying and dangerous to see the right to an explanation become a similarly empty formality. This ``transparency fallacy'' is something we should guard against and that should spur us on, as in the next section, to look at alternative and supplementary ways to build better systems.

In particular, any remedy given by a right to an explanation will often come too late for that specific user. The current investigations into automated recidivism decisions in the US biased against black data subjects show very clearly a history of discrimination for years, which is only now becoming apparent. In many cases, we would rather the system was never broken in first place, or at very least that the individual decision concerned swung in a more just direction. This takes us to a range of ex ante as well as ex post governance tools.

In the next sections, we begin to consider what other regulatory tools have been created in the GDPR and elsewhere that might be pressed into use to try to ensure, audit, or instigate the creation of algorithms that are fairer, less discriminatory, and ideally, less opaque.

\section{Investigating before Deployment or Decision-Making}\label{investigating-before-deployment-or-decision-making}

When algorithmic systems make choices in real time, such as who can access a service or who is selected for security screening, redress or transparency long after the choice is little help. For this, we must consider the governance tools that have impacts upstream, while systems are being designed or at least before they are deployed.

\subsection{Privacy by Design, Data Protection by Design, and Impact Assessments}\label{privacy-by-design-data-protection-by-design-and-impact-assessments}

The GDPR introduces a number of new provisions that, rather radically, do not confer individual rights, but rather attempt to create an environment in which less ``toxic'' automated systems will be built in future. These ideas come out of the long evolution of privacy by design (PbD) engineering as a way to build privacy-aware or privacy-friendly systems, generally in a voluntary rather than mandated way. They recognize that a regulator cannot do everything by top-down control, but that controllers must themselves be involved in the design of less privacy-invasive systems. These provisions include the following requirements:

\begin{itemize}
\item  Controllers must, at the time systems are developed as well as at the time of actual processing, implement ``appropriate technical and organisational measures'' to protect the rights of data subjects (GDPR, article 25). In particular, ``data protection by design/default'' is required so that only personal data necessary for processing is gathered. Suggestions for PbD include making use of pseudonymization and data minimization.
\item  When a type of processing using new technologies is ``likely to result in a high risk'' to the rights of data subjects, then there must be a prior data protection impact assessment (DPIA) (article 35), and under some conditions, consultation with the regulator.
\item Every public authority, every ``large scale'' private-sector controller, and any controller who processes the special categories of data under article 9 (sensitive personal data) must appoint a data protection officer (DPO) (article 37).
\end{itemize}

DPIAs especially have tremendous implications for ML design. Impact assessments are tools used in many domains to assess or estimate impacts of particular interventions or courses of action. GDPR article 35 notes that:

Where a type of processing in particular using new technologies \ldots{} is likely to result in a high risk to the rights and freedoms of natural persons, the controller shall, prior to the processing, carry out an assessment of the impact of the envisaged processing operations on the protection of personal data.

Where a DPIA ``indicates that the processing would result in a high risk in the absence of measures taken by the controller to mitigate the risk,'' the controller is obliged to ``consult {[}the data protection authority{]} prior to processing'' (article 36(1)) with a view to putting in measurers to mitigate the risk. Realistically, this seems only likely in cases of highly novel technologies or the use of existing technology in a new context; DPIAs are not intended as a tool to stop processing, but rather as a way to refine, or provide points of accountability for the future operation of, complex systems.

The Article 29 Working Party (A29 WP; a body made up of national DP supervisory authorities that gives authoritative but not binding recommendations on how to interpret DP laws---soon to be revamped as the European Data Protection Board) has issued draft guidance (17/EN WP 248) further elaborating the conditions under which a data controller must carry out a DPIA. These include if two or more of the following conditions are met as part of processing:

\begin{itemize}
\item evaluation or scoring of individuals;
\item automated decision making with legal or similar significant effect;
\item systematic monitoring, including of a public area;
\item sensitive data processing, as defined in the GDPR;
\item data processed on a large scale;
\item matched or combined datasets, particularly if data subjects might have had different expectations about their use;
\item data concerning vulnerable data subjects;
\item innovative use of technological or organizational solutions;
\item data transfer outside the EU; and
\item processing that prevents rights being exercised, such as in a public area people cannot avoid, or as a necessary prerequisite to service provision.
\end{itemize}

Taken together with the GDPR, this guidance indicates that a DPIA will be an obligatory precursor for many ML systems with sizable anticipated risks or consequences for individuals or groups.

DPIAs are not a replacement for explanations of algorithmic systems, not least because they are aimed at helping builders and regulators, not, directly, users (although article 35(9) does provide that ``where appropriate, the controller shall seek the views of data subjects''). DPIAs are also not required to be public documents, although it is considered good practice by the A29 WP to do so at least in part. However, they may be of considerable value in leading to the building of better systems overall.

Articles 35 and 36 do not specifically require a DPIA to combat potential discrimination: early drafts made a DPIA mandatory where there was a ``risk of discrimination being embedded in or reinforced by the operation.''\textsuperscript{10} This amendment in the final text was relegated to recitals (see recitals 71 and 75), which as discussed, have a murky status in EU law. However, in its draft guidance, the A29 WP explicitly clarified that ``rights and freedoms'' in article 35(1) ``may also involve other fundamental rights such as \ldots{} prohibition of discrimination.'' The UK's Information Commissioner's Office (ICO), in its influential guidance report on big data, artificial intelligence, machine learning, and data protection,\textsuperscript{11} notes firmly that ``potential privacy risks'' have already been identified with ``the use of inferred data and predictive analytics'' and goes on to provide a draft DPIA for big data analytics (Annex 1). It seems clear that, despite the uncertainty of the ``high risk'' threshold, DPIAs are quite likely to become the required norm for algorithmic systems, especially where sensitive personal data, such as race or political opinion, is processed on a ``large scale'' (GDPR, article 35(3)(b)).

Impact assessments that deal with risks of discrimination do already exist. The UK has extensive experience with equality impact assessments (EqIAs), which used to be required for every new governmental policy and were considered one of the main ways of documenting fulfilment of the Public Sector Equality Duty, which was brought into law by the Equality Act 2010. The Public Sector Equality Duty requires due regard to be given to impacts on protected classes before a policy is finalized or implemented, and it is primarily accessed through judicial review. This requirement does not always need to be met via an EqIA (see \emph{R (Brown) v Sec of State for Work and Pensions} 2008 EWHC 3158 (Admin)) but must be carried out with rigor before a policy is implemented, with documentation to show the process if it is not otherwise clear. Arguably, where new public sector decision support systems are built, an EqIA would be highly appropriate and could be combined with the DPIA process. Recent proposals for ``algorithmic impact assessments'', while an immature field awaiting concrete proposals, would appear to straddle EqIAs and DPIAs in form.

\subsection{Certification Systems}\label{certification-systems}

The GDPR also introduces the idea of voluntary certification for ML systems. Article 42 proposes voluntary certification of controllers and processors to demonstrate compliance with the regulation, with ``certification mechanisms'' and the development of ``seals and marks'' to be encouraged by EU member states. In the UK, a tender has already been advertised by the ICO for a certification authority to run a UK privacy seal, although progress has been interrupted by the vote to exit the European Union and the subsequent political turmoil.

Taken together, these provisions offer exciting opportunities to operationalize what in the US have been called ``big data due process'' rights.\textsuperscript{12,13} Certification could be applied to two main aspects of algorithmic systems:

\begin{itemize}
\item certification of the algorithm as a software object by (a) directly specifying either its design specifications or the process of its design, such as the expertise involved (technology-based standards) and/or (b) specifying output-related requirements that can be monitored and evaluated (performance-based standards); and
\item certification of the whole person or process using the system (system controller) to make decisions, which would consider algorithms as situated in the context of their use.
\end{itemize}

One notable advantage is that certification standards could be set on a per-sector basis. This is already very common in other sociotechnical areas, such as environmental sustainability standards.

The downside of what seems an exciting approach is that the history in the privacy domain of self-regulation of the private sector by seal and certificates is dispiriting. Essentially this involves privatization of regulation and scrutiny. Certification scheme and trust seals have to make money to survive, which can only be obtained by asking fees from members. Given this self-interest, it is hard to punish members too hard when they breach the rules of the seal or certificate, for fear they will leave, either altogether or for a less demanding trust seal (in a plural market, which is generally what is envisaged). This in turn tends to diminish the value of the seal or certificate as a guarantee of trustworthiness. There is also little proof users regard seals and certificates as indicators of trust, which may mean organizations are unwilling to pay or make an effort to belong to them, unless by doing so they can avoid more stringent top-down regulation.

A clear example here is the US--EU Safe Harbor agreement for the export of personal data to the US, which came ignominiously to an end in \emph{Schrems v DPC} of Ireland in the Court of Justice of the EU in 2015 (Case C-362/14). The US as a country had been deemed not to have adequate protection in its law for personal data, and so in principal, EU data could not be exported to the US as far back as the DPD in 1995. A solution was found, however, in the Safe Harbor agreement whereby US companies could receive EU data if they joined a trust seal, membership of which guaranteed they were meeting adequate privacy standards. One of the largest and most prominent seals used in this way was TrustE. Yet as Charlesworth showed back in 2000,\textsuperscript{14} TrustE had a long history of overlooking major data breaches by members or imposing only desultory sanctions. This, as well as the US's cavalier attitude to covert state surveillance exposed in the Snowden revelations, led to the agreement's judicial annulling.

\section{Enabling Review and Challenge Without Individual Burden}\label{enabling-review-and-challenge-without-individual-burden}

Injustices in decision-making can also emerge even from algorithmic systems thought to be well designed, and in those cases, reactive provisions are appropriate and needed. But how can these be achieved whilst avoiding burdening individuals with the responsibility of understanding, investigating, and identifying points of action in arcane and often shadowy technological deployments?

\subsection{Representation Bodies for Data Subjects}\label{representation-bodies-for-data-subjects}

The aim of a right to an explanation is fundamentally to enable challenge to poor or wongly made decisions by black box systems. Yet, as we have seen, individual users typically The to assert their rights, especially in complex and opaque areas such as ML systems. Moreover, if we are talking about a harm such as embedded systemic discrimination, which typically affects a whole class of people, then it may seem far more appropriate for a representative body to accept and mount challenges than each individual user. This is a common problem in consumer law, and was long anticipated in DP law by the creation of state DPAs, which already have a role to investigate user complaints and enforce breaches against data controllers. One historic problem here has been the low level of sanctions a DPA could dish out: this has famously been met in the GDPR by the creation of a maximum fine of up to EUR 20 million or 4 percent of global annual turnover. This welcome change does not alter the fact, however, that the volume of DP breaches both deliberate and inadvertent is now so huge that one agency alone cannot combat it. Furthermore, state DPAs throughout the EU are wildly underfunded, since in their nature (and by law), they have to be seen to be independent of both state and the commercial sector. A final major problem germane to ML-related complaints is that DPAs are typically staffed by civil servants and/or lawyers and rarely have much technical understanding or capacity.

The GDPR tries to help here in two main ways. Article 80(1) provides for all member states that a data subject can mandate a third body to lodge a complaint, exercise the right to judicial remedy, and receive compensation on his or her behalf. This is a useful step, particularly if civil society bodies can find an exemplary case to support, but it still requires a data subject to notice a breach and have the time and effort to reach out to a third body and enlist its help. Given a third body can receive compensation on behalf of a data subject, this might also result in a dubious ``ambulance-chasing'' industry (similar to the UK furor over wrongly paid payment protection insurance, or PPI).

Article 80(2), GDPR, by contrast, permits member states to allow third-party bodies to take up complaints, for example, against a data controller, without being mandated by a data subject. Through this provision, civil society bodies could monitor sectors and controllers for breaches and pursue suspected infringements of their own accord through judicial remedies.

Article 80(1) is mandatory for GDPR-implementing states to put into law, whereas article 80(2) is not. While some countries such as Germany will implement this, in the UK the Government seems unwilling to, having only conceded to review the functionality after significant parliamentary pressure. This seems odd, given the UK has already nominated ``super-complaint'' non-governmental organizations (NGOs) that can, for example, raise consumer or financial rights issues to regulators on behalf of groups they perceive as affected.

Bodies like this might become effective watchdogs in particular areas or sectors, but they will find it difficult to do this without having the capability for some access in the round to training data, input data, outputs, and models of algorithmic systems in order to establish whether breaches are occurring. A large problem here is that courts have typically been reluctant to order access to source code for decision-making systems even in relation to traditional non-algorithmic systems. This is of course partly because of the issue of proprietary IP rights in code already noted above (see, for instance, \emph{Viacom v YouTube} Civ. 2103 (LL) in the US), but other issues may also be implicated.

In the UK, there appears to be no reported case where a court has ordered disclosure of the source code of a decision support system to litigants, even in the surprisingly high number of disputes involving public sector systems, where issues of copyright might have been thought to be less prominent. It is interesting though that in at least one case, \emph{Northern Metco Estates Ltd v Perth \& Kinross District Council} 1993 S.L.T. (Lands Tribunal) 28, the output of a conventional though complex automated decision support system was doubted in respect to its value without more information as to how it was generated. The system in question calculated one factor (economic rent) to feed into a compensation valuation in cases of compulsory acquisition of land. The court was disturbed at the lack of evidence it received as to exactly how this calculation had been done but, interestingly, did not seem interested to find out more but rather to exclude it from influence. It seems quite likely that courts will be reluctant to become activists about disclosures of source code, let alone algorithmic training sets and models, until they feel more confident of their ability to comprehend and use such evidence---which may take some time.

This likely judicial reluctance to become involved in unravelling algorithmic systems is unfortunate because, at least in the public sector, a very valuable tool for transparency might be found in the institution of judicial review. In the UK, courts have the power on petition to review if administrative acts and discretion were legal and reasonable. In principle, there seems no reason why the court's powers in such review might not extend to demanding access to data and models. However, as noted above, there is as yet no record of this having been done. This may be due to a combination of proprietary code issues and technophobia, but there may also be worries about exposing sensitive personal data to the public eye (court documents are usually public).

One way forward here may be drawn from developing practice around sensitive public sector data and access to it by non-governmental data analysts. In recent years, much work has been put into building methods by which sensitive data can be accessed and analyzed securely, proportionally, and safely. National statistics offices have been working on ways to enable access to these rich troves of information by external data analysts without compromising security and privacy. Although several methods have been used here, the most interesting for our purposes is the development of secure environments for access to sensitive data. These locations are usually disconnected from a network and may be physically located in a public agency, private company, or increasingly, academia.

We imagine that such systems might scale to become depositories for the scrutiny of the models and data associated with public sector algorithmic systems. Challenges to private sector systems might also benefit from such systems, albeit with IP issues to overcome. Archival or specialist libraries may yet become homes for infrastructure to publicly scrutinize models and code, just as they already often act as code escrow agencies.\\

As things stand, the right to an explanation found in the GDPR, though beguiling, is uncertain, convoluted, rife with technical difficulties, and likely to be interpreted differently in different member states. There are key restrictions in the remedies it gives, which are likely to exclude the decision support systems where transparency would be most welcomed: in crucial sectors such as criminal justice, policing, and child protection where decisions are not solely based on automated processing. As a rights-based remedy, it places a large burden on individual users to not only seek their own explanations but follow up with challenges. It also does not work well as a remedy for harms experienced in aggregate by groups or protected classes. There is a danger that the hype around the right to an explanation may create the belief that there is no need for further legal and non-legal solutions. Some newer legal instruments, such as the new French law, offer refinements to the GDPR but may also add their own restrictions.

Accordingly, we proposed a wider sweep of attention toward other legal and paralegal remedies that may also have scope to impel the creation of better and more scrutable algorithmic systems. These include privacy by design, data protection impact assessments, and certification or seal schemes, all of which might offer scrutiny during the process of building systems rather than merely after they cause harm. We also suggested that redress for users after a system has caused harm could be improved by taking the opportunities the GDPR gives for representation and collective action by non-governmental bodies. Finally, we noted that access to source code and data of algorithms will remain a practical issue for NGOs as well as in contested legal cases for as long as courts are reluctant to order its disclosure, and suggested secure model depositories might help in this area.

\section*{Acknowledgments}

Lilian Edwards was supported in part by the Arts and Humanities Research Council (AHRC) Centre CREATe (grant number AH/K000179/1), and the EPSRC Digital Economy Hub Horizon at the University of Nottingham (grant number EP/ G065802/1). Michael Veale acknowledges funding from the Engineering and Physical Sciences Research Council (EPSRC) (grant number EP/ M507970/1).

\section*{References}
\small
\begin{enumerate}
\def\labelenumi{\arabic{enumi}.}
\item
  L. Edwards and M. Veale, ``Slave to the Algorithm? Why a ``Right to an Explanation'' Is Probably Not the Remedy You Are Looking For,'' \emph{Duke Law \& Technology Review}, vol. 16, 2017, pp. 18--84, doi:10.2139/ssrn.2972855.
\item
  F. Pasquale, \emph{The Black Box Society: The Secret Algorithms That Control Money and Information}, Harvard University Press, 2015.
\item
  M. Veale and L. Edwards, ``Clarity, Surprises, and Further Questions in the Article 29 Working Party Draft Guidance on Automated Decision-Making and Profiling,'' \emph{Computer Law \& Security Review}, vol. 34, no. 2, 2018, doi:10.1016/j.clsr.2017.12.002.
\item
  L. Sweeney, ``Discrimination in Online Ad Delivery,'' \emph{Queue}, vol. 11, no. 3, 2013, p. 10.
\item
  I. Mendoza and L.A. Bygrave, ``The Right Not to Be Subject to Automated Decisions Based on Profiling,'' \emph{EU Internet Law: Regulation and Enforcement}, T.-E. Synodinou et al., eds., Springer, 2017.
\item
  S. Wachter et al., ``Why a Right to Explanation of Automated Decision-Making Does Not Exist in the General Data Protection Regulation,'' \emph{International Data Privacy Law}, vol. 7, no. 2, 2017, pp. 76--99, doi:10.1093/idpl/ipx005.
\item
  M.T. Ribeiro et al., ``Why Should I Trust You?: Explaining the Predictions of Any Classifier,'' \emph{Proceedings of the 22nd ACM SIGKDD International Conference on Knowledge Discovery and Data Mining}, 2016, pp. 1135--1144.
\item
  A.B. Tickle et al., ``The Truth Will Come to Light: Directions and Challenges in Extracting the Knowledge Embedded within Trained Artificial Neural Networks,'' \emph{IEEE Trans. Neural Netw}., vol. 9, no. 6, 1998, pp. 1057--1068.
\item
  G. Montavon et al., ``Explaining Nonlinear Classification Decisions with Deep Taylor Decomposition,'' \emph{Pattern Recognit}., vol. 65, 2017, pp. 211--222.
\item
  R. Binns, ``Data Protection Impact Assessments: A Meta-Regulatory Approach,'' \emph{International Data Privacy Law}, vol. 7, no. 1, 2017, pp. 22--35, doi:10.1093/idpl/ipw027.
\item
  ``Big Data, Artificial Intelligence, Machine Learning and Data Protection,'' Information Commissioner's Office, 2017.
\item
  D.K. Citron, ``Technological Due Process,'' \emph{Washington University Law Review}, vol. 85, 2008, pp. 1249--1313.
\item
  K. Crawford and J. Schultz, ``Big Data and Due Process: Toward a Framework to Redress Predictive Privacy Harms,'' \emph{Boston College Law Review}, vol. 55, 2014, pp. 93--128.
\item
  A. Charlesworth, ``Clash of the Data Titans---US and EU Data Privacy Regulation,'' \emph{Eur. Pub. L.}, vol. 6, 2000, p. 253.
\end{enumerate}

\printbibliography
\end{document}